\documentclass[sigconf, nonacm]{acmart}

\AtBeginDocument{%
  \providecommand\BibTeX{{%
    \normalfont B\kern-0.5em{\scshape i\kern-0.25em b}\kern-0.8em\TeX}}}

\usepackage{xspace}
\usepackage{siunitx}
\usepackage{mathtools}

\usepackage{hhline}
\usepackage{multirow}
\usepackage{diagbox}
\usepackage{tikz} 
\usetikzlibrary{decorations.pathreplacing}
\usepackage{threeparttable, tablefootnote}
\usepackage{placeins}
\newcommand\users{\ensuremath{U}\xspace}
\newcommand\items{\ensuremath{I}\xspace}
\newcommand\useru{\ensuremath{u}\xspace}
\newcommand\usentences{\ensuremath{S_{\useru}}\xspace}
\newcommand\isentences{\ensuremath{S_{\itemi}}\xspace}
\newcommand\uisentences{\ensuremath{S_{\useru\itemi}}\xspace}
\newcommand\visentences{\ensuremath{S_{\userv\itemi}}\xspace}
\newcommand\userv{\ensuremath{v}\xspace}

\newcommand\vsentences{\ensuremath{S_{\userv}}\xspace}
\newcommand\itemi{\ensuremath{i}\xspace}
\newcommand\itemj{\ensuremath{j}\xspace}

\newcommand\sentencegraph{\ensuremath{G}\xspace}
\newcommand\vertices{\ensuremath{V}\xspace}
\newcommand\edges{\ensuremath{E}\xspace}
\newcommand\edgeweight{\ensuremath{s\left(\sentenceu, \sentencev\right)}\xspace}

\newcommand\globalgraph{\ensuremath{G}\xspace}

\newcommand\usersim{\ensuremath{w\left(\useru,\userv\right)}\xspace}

\newcommand\graphusersimoneone{\ensuremath{w\left(\useru,\userv\right)}\xspace}
\newcommand\graphusersimmanyone{\ensuremath{w\left(\useru,\userv\right)}\xspace}
\newcommand\graphusersimmanymany{\ensuremath{w\left(\useru,\userv\right)}\xspace}

\newcommand\cosinesim{\ensuremath{\text{cosine similarity}}\xspace}

\newcommand\sentenceu{\ensuremath{\sigma_1}\xspace}
\newcommand\sentencev{\ensuremath{\sigma_2}\xspace}
\newcommand\neighborhood{\ensuremath{N}\xspace}
\newcommand\userneighborhooduser{\ensuremath{\neighborhood\left(\useru\right)}\xspace}
\newcommand\userneighborhoodsentence{\ensuremath{\neighborhood\left(\sentenceu\right)}\xspace}

\newcommand\polarization{\ensuremath{p\left(\sentenceu, \sentencev\right)}\xspace}
\newcommand\sentiment{\ensuremath{\alpha}\xspace}

\newcommand\lfcp{\ensuremath{\text{TFCP}}\xspace}

\newcommand\lfcpmacro{\ensuremath{\text{TFCP}}\xspace}
\newcommand\weightedindegree{\ensuremath{\delta^-}\xspace}  
\newcommand\weightedoutdegree{\ensuremath{\delta^+}\xspace}
\newcommand\indicator{\ensuremath{I\!}\xspace}

\begin{document}

\title{KNNs of Semantic Encodings for Rating Prediction}

\author{Léo Laugier}
\authornote{Work done at Institut Polytechnique de Paris.}
\orcid{0000-0002-3737-3092}
\affiliation{%
  \institution{EPFL}
  \streetaddress{Route Cantonale}
  \city{Lausanne}
  \country{Switzerland}
  \postcode{1015}
}
\email{leo.laugier@epfl.ch}

\author{Raghuram Vadapalli}
\affiliation{%
  \institution{Google}
  \streetaddress{19 place Marguerite Perey}
  \city{London}
  \country{United Kingdom}}
\email{rvadapalli@google.com}

\author{Thomas Bonald}
\affiliation{%
  \institution{Télécom Paris, Institut Polytechnique de Paris}
  \streetaddress{19 place Marguerite Perey}
  \city{Palaiseau}
  \country{France}}
\email{thomas.bonald@telecom-paris.fr}

\author{Lucas Dixon}
\affiliation{%
  \institution{Google}
  \streetaddress{19 place Marguerite Perey}
  \city{Paris}
  \country{France}}
\email{ldixon@google.com}







\begin{abstract}
  This paper explores a novel application of textual semantic similarity to user-preference representation for rating prediction. The approach represents a user's preferences as a graph of textual snippets from review text, where the edges are defined by semantic similarity. This textual, memory-based approach to rating prediction enables review-based explanations for recommendations. The method is evaluated quantitatively, highlighting that leveraging text in this way outperforms both strong memory-based and model-based collaborative filtering baselines. 
\end{abstract}






\maketitle

\section{Introduction}

With the democratization of online shopping and content delivery platforms, people have become accustomed to evaluating items online. For example, before watching a movie, it is common to look up the movie in a search engine and read more about it. To support this, search engines and movie streaming sites, like Netflix, show information to help such as the average rating, public reviews, personalized ratings, and snippets of text that explain why the user may enjoy it, so-called endorsements.

\begin{figure}
    \centering
    \includegraphics[width=0.45\textwidth]{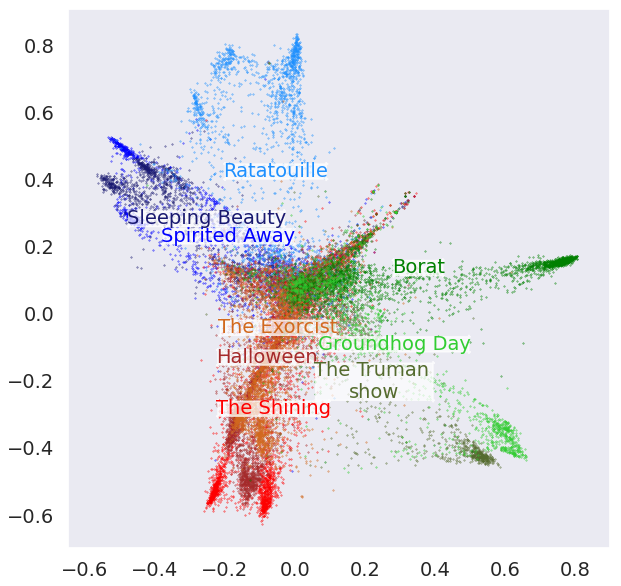}
    \caption{2D t-SNE projection of sentence embeddings. Red, green, and blue colors represent sentences from horror, comedy, and animation movies, respectively. Movie titles are displayed at the barycenter of their review sentences (with adjustment for readability purposes). Figure best viewed in color.}
    \Description{Fully described in the text.}
    \label{fig:tsne}
\end{figure}

This paper focuses on rating prediction: the task of predicting the rating a person will give an item. This is particularly relevant when someone is evaluating whether to spend time or money on a given item. Statistical and machine learning models have often been used to compute rating predictions based on past interactions, and the archetypal baseline for such systems is Collaborative Filtering (CF) \citep{Ricci2011}.
While rating prediction systems initially relied only on quantitative preference scores, newer datasets with textual reviews combined with the recent advances in natural language understanding have opened up the research directions that explore possible benefits for both explainability and performance of rating prediction systems. For example, \citet{info11060317} have summarised research on Collaborative Filtering (CF) that leverages review text. They follow a classification of systems outlined by \citet{10.1007/s11257-015-9155-5} and note benefits when ratings are sparse, and performance improvements to rating-prediction. However, these studies largely explore high-level derived features such as sentiment or topic-clusters; far less research has explored the use of semantic information from unstructured text from reviews.

Memory-based and model-based are the two main families of methods in CF \citep{aggarwal2016recommender, 10.5555/2770959.2771069}. Memory-based CF predicts user-ratings by directly aggregating ratings from other users with similar preferences. Model-based CF uses interaction data to train a user and item model - typically high dimensional vectors - that can be used to generate ratings for future items. Even when memory-based CF shows worse absolute quantitative evaluation on standard metrics, it benefits from the simplicity of implementation and ease of creating explanations \citep{8506344}. 

\begin{figure*}[!ht]
    \centering
    \begin{tikzpicture}[	
		layer/.style={
			rectangle,
			minimum size=6mm,
			thick,
			top color=white,
			font=\itshape
			}, bullet/.style={fill,circle,inner sep=2pt}, empty/.style={shape=circle,inner sep=2pt, draw},
			squarebullet/.style={fill,rectangle,inner sep=2pt}, squareempty/.style={shape=rectangle,inner sep=2pt, draw},]	
			
	    \node[white] (sentence_text_fantom) at (0,0) {It was wonderfully funny.};
		\node[black] (sentence_text) at (0,0.7) {a great family movie.};
		\node[black] (sentence_textp) at (0,.2) {It was wonderfully funny.};
		\node[black] (sentence_text_3) at (0,-0.7) {A horror movie classic .};
		\draw[color=blue!60!green, dashed, rounded corners] (-2,0) rectangle (2,1);
		\draw[red, dashed, rounded corners] (-2,-1) rectangle (2,-0.5);

		
		\path (sentence_textp) -- (sentence_text_3) node [black, midway, sloped] {$\dots$};	
		\node[layer, draw=blue!100!black!50, bottom color=blue!100!black!20] (use) at (3,0) {USE};
		\node[white] (sentence_emb_fantom) at (4.2,0) {$\sigma_1''$};
		\node[black] (sentence_emb) at (4.2,1.5) {$\sigma^{\hphantom{''}}$};
		\node[black] (sentence_embp) at (4.2,1) {$\sigma'^{\hphantom{'}}$};
		\node[black] (sentence_emb_1) at (4.2,0.5) {$\sigma_1^{\hphantom{''}}$};
		\node[black] (sentence_emb_1p) at (4.2,0) {$\sigma_1'^{\hphantom{'}}$};
		\node[black] (sentence_emb_1pp) at (4.2,-0.5) {$\sigma_1''$};
		\node[black] (sentence_emb_2) at (4.2,-1) {$\sigma_2^{\hphantom{''}}$};
		\node[black] (sentence_emb_3) at (4.2,-1.5) {$\sigma_3^{\hphantom{''}}$};
		\draw[color=blue!60!green, dashed, rounded corners] (3.9,0.75) rectangle (4.5,1.7);
		\node[white] (user_phantom_bis) at (4.8,0) {$v_1$};
		\node[black] at (4.8,1.25) {$u_{\hphantom{1}}$};
		\draw[color=blue!80!green, dashed, rounded corners] (3.9,-0.75) rectangle (4.5,0.7);
		\node[black] at (4.8,0) {$v_1$};
		\draw[color=blue!40!green, dashed, rounded corners] (3.9,-1.25) rectangle (4.5,-0.8);
		\node[black] at (4.8,-1) {$v_2$};
		\draw[red, dashed, rounded corners] (3.9,-1.75) rectangle (4.5,-1.3);
		\node[black] at (4.8,-1.5) {$v_3$};

        \begin{scope}[shift={(6.7,1)}]
            \node[label=above:Sentence k-NN graph] (rect1) at (1,-1) [draw,rounded corners,minimum width=4cm,minimum height=4cm] {};
        
            \begin{scope}[nodes=bullet]
               \node[color=blue!60!green, label=left:$\sigma$] (sigma) at (0,0) {};
               \node[color=blue!60!green, label=right:$\sigma'$] (sigmap) at (0.7,-2) {};
            \end{scope}
            
            \begin{scope}[nodes=empty]
               \node[color=blue!80!green, label=right:$\sigma_{1}$] (sigma1) at (0.5,0.5) {};
               \node[color=blue!40!green, label=right:$\sigma_{2}$] (sigma2) at (1,-0.5) {};
               \node[red, label=right:$\sigma_{3}$] (sigma3) at (2,-1) {};
               
               \node[color=blue!80!green, label=left:$\sigma_{1}'$] (sigma1p) at (0.2,-1.5) {};
               \node[color=blue!80!green, label=left:$\sigma_{1}''$] (sigma1pp) at (0,-2.5) {};
            \end{scope}
            
            \begin{scope}
               \path [->] (sigma) edge node[midway,above,sloped]{$0.9$} (sigma1) ;
               \path [->] (sigma) edge node[midway,below,sloped]{$0.7$} (sigma2);
               
               \path [->] (sigmap) edge node[midway,above,sloped]{$0.9$} (sigma1p) ;
               \path [->] (sigmap) edge node[midway,below,sloped]{$0.8$} (sigma1pp);
               
            \end{scope}
		\end{scope}

		\begin{scope}[shift={(11.7,0.3)}]
            \node[label=above:User k-NN regressor] (rect2) at (0.75,-0.25) [draw,rounded corners,minimum width=4cm,minimum height=2.5cm] {};
        
            \begin{scope}[nodes=squarebullet]
               \node[color=blue!60!green, label=right:$u$] (u) at (0,0) {};
            \end{scope}
            
            \begin{scope}[nodes=squareempty]
               \node[color=blue!80!green, label=left:$v_{1}$] (v1) at (-0.5,0.5) {};
               \node[color=blue!40!green, label=right:$v_{2}$] (v2) at (1,-1) {};
               \node[red, label=right:$v_{3}$] (v3) at (2,0) {};
            \end{scope}
            
            \begin{scope}
               \path [-] (u) edge node[near end,above,sloped]{$3$} (v1) ;
               \path [-] (u) edge node[near end,above,sloped]{$1$} (v2);
            \end{scope}
		\end{scope}
		
		\node[black] (prediciton) at (12.5,-2) {$\hat{r}_{uj}=\frac{3}{4}\cdot r_{v_1j} + \frac{1}{4} \cdot r_{v_2j}$};
		
		\draw[->,>=latex,thick] (sentence_text_fantom) -- (use);
		\draw[->,>=latex,thick] (use) -- (sentence_emb_fantom);	
		\draw[->,>=latex,thick] (user_phantom_bis) -- node[above] {$s$} (rect1);
		\draw[->,>=latex,thick] (rect1) -- node[above] {$w$} (rect2);
		
		\draw[->,>=latex,thick] (rect2) -- (prediciton);
		
	\end{tikzpicture}
    \caption{Text-KNN pipeline. $\sigma$ and $\sigma'$ are the sentences written by user \useru $\in$ \users. $r_{vj}$ is the rating of user \userv $\in$ \users on item \itemj $\in$ \items. $\sigma_n, \sigma_n', \sigma_n'', \hdots$  represent the embeddings of sentences written by user $v_n$. k-NN graphs are built from the sentence similarity $s$, and are used for computing the weights $w$ of the user k-NN regressor. Figure best viewed in color.}
    \Description{Fully described in the text.}
    \label{fig:pipeline}
\end{figure*}
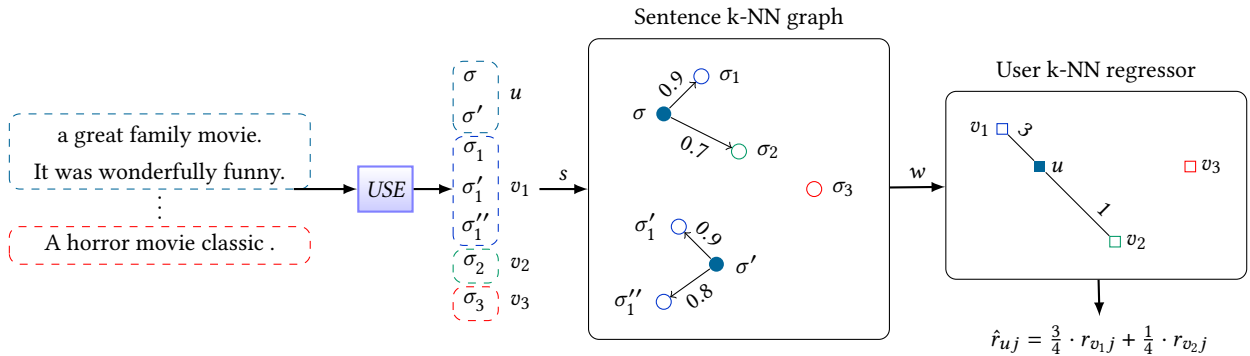

In this paper, we introduce a novel memory-based approach to rating-prediction. It leverages review text by using recent advances in pre-trained models of textual semantic similarity coming from machine learning research applied to Natural Language Processing (NLP). The new approach also introduces a new way to support explanations for recommendations based on snippets of review text. In particular, we use modern large pre-trained language models \citep{yang-etal-2020-multilingual} to embed sentences of reviews into a so called semantic vector space (Section~\ref{sec:semantic_match}). We then employ a user k-NN model based on user similarity (Section~\ref{sec:rec_sys}). The key insight is that user similarity can be defined in terms of a graph based on semantic similarity from the text in user-reviews. Specifically, we show how to build sentence-based k-nearest neighbors (k-NN) graphs (Section~\ref{sec:s}) to compute user similarity. We denote our model Text-KNN\footnote{Our code is available at \url{https://bit.ly/3NlL6Qa}} (Figure~\ref{fig:pipeline}).





We quantitatively evaluate our model in the context of two large datasets containing both numerical and text reviews; the Amazon Review dataset \citep{10.1145/2872427.2883037} and the Yelp dataset \citep{yelp}. To avoid the problems frequently highlighted with RMSE-based evaluation \citep{10.1145/1864708.1864721}, we follow the approach of \citet{Koren2013CollaborativeFO}.\footnote{It should be noted that top-N ranking metrics, such as MAP@k, Hit-Rate and DCG \citep{10.1145/582415.582418}, while effective for scoring ranking systems such as next-item recommenders, do not apply to rating prediction.} The evaluation highlights that our proposed KNN model beats strong baselines for both memory-based and model-based systems. The result is that our model provides both explainability benefits, inherited from memory-based methods, enhanced by now enabling textual-review snippets to be used, as well as competitive performance.

%


\section{Related work} \label{sec:related}

User k-NN is the main approach studied in memory-based CF. The cornerstone of k-NN is the definition of a similarity measure between two users \citep{Ning2015}. Several similarities can be computed from the vectors of the ratings of co-reviewed items, among which mean squared difference (MSD, \citealp{10.1145/223904.223931, Pazzani2004AFF}), cosine similarity \citep{1167344}, Pearson correlation coefficient \citep{10.1145/963770.963772} and Jaccard coefficient \citep{10.1145/1559845.1559923}.

The use of textual reviews for memory-based CF has received relatively little attention \citep{Leung2006IntegratingCF, 10.1145/2365952.2365977, 10.5555/2540128.2540515}. To the best of our knowledge, the only previous effort  to leverage text in memory-based CF is that of \citet{3195c13d3d90462f8aa02b15d6e7d35d}. They explored user-based k-NN similarities based on the {\em words} found in text reviews: they introduce six word-based similarity measures ranging from simple word overlap, to depth of two words \citep{09c1a5ee793d4263a05e572380f6b3ab} in external lexical graphs\footnote{WordNet was employed \citep{miller1998wordnet}.}, and information content \citep{10.5555/1625855.1625914, jiang-conrath-1997-semantic, 10.5555/645527.657297}. Recent progress in Natural Language Understanding (NLU) enables far richer and more effective semantic representations of text \citep{NIPS2017_3f5ee243, devlin-etal-2019-bert, 2020t5}. While this work is similar in spirit \citet{3195c13d3d90462f8aa02b15d6e7d35d}, it employs these recent advances in language models of semantic similarity which operate at the sentence level. The result is that our rating prediction model has significantly stronger performance.

Model-based CF relies on intrinsic models trained with various machine learning techniques \citep{Su2006CollaborativeFF, 10.1145/1185448.1185487, 10.1145/3285029} to directly predict user preferences. In particular, state-of-the-art latent factor models \citep{5197422} such as probabilistic matrix factorization \citep{salakhutdinov2008a}, Singular Value Decomposition (SVD) \citep{Funk} and SVD++ \citep{10.1145/1644873.1644874} achieve the most precise rating predictions.
There has been a large body of work extending model-based methods that were initially designed to use only the history of user-ratings. Examples of additional sources of information include social information \citep{su2017effective}, tags \citep{info9060143, 10.1145/3331184.3331211} and item descriptions \citep{Alshammari, DBLP:journals/corr/abs-2109-09358}.
Several authors have argued for the effectiveness of using the text found in reviews \citep{Zhang, 10.1007/s11257-015-9155-5, hernandez, 10.1145/3477495.3531873}. Strategies for doing so have focused on creating topic-cluster representations \citep{10.1145/2507157.2507163, Bao_Fang_Zhang_2014, 10.1145/3269206.3271810} and using sentiment scores \citep{poirier:inria-00514533, 10.1145/2600428.2609579, Dau}. 
Another strategy for model-based methods has been to build latent representations of words or sentences with neural networks, before integrating these vector-based representations into matrix factorization CF \citep{10.1145/2959100.2959165, 10.1145/3018661.3018665, 10.1145/3178876.3186070, Liu2020HybridNR}. However, for memory-based methods, there has been very little exploration. Our contribution explores semantic similarity encoding of sentences for a memory-based model.


A large body of research closely related to item prediction concerns next item recommendation and information retrieval (IR) systems \citep{buttcher2016information}. These have evolved to use different metrics, notably top-N ranking metrics, such as discounted cumulative gain \citep{10.1145/582415.582418} and hit rate. Rating prediction has traditionally been evaluated with error-based metrics, notably root mean squared error (RMSE). However, RMSE is not effective at distinguishing the quality of algorithms, and high performance can be achieved by naive popularity baselines~\cite{10.1145/1864708.1864721}. Moreover, users who employ a smaller spread of scores are less represented by RMSE (as error values for them will be smaller). To remedy this, we report a metric based on the Fraction of Concordant Pairs (FCP) \citep{Koren2013CollaborativeFO}. More details can be found in Section~\ref{sec:evaluation}. 

One interesting connection between our work, IR endorsements \citep{DBLP:conf/sigir/LiLS11} and snippet extraction \citep{10.1016/j.patcog.2009.06.003} is that our method provides a natural way to create a similar kind of ``explanation''. Our method produces sentences from reviews based on their role in a user's predicted rating.

\section{Explicit feedback has less confounds}
Preferences are only partially reflected by the numerical scores in reviews. Similarly, implicit feedback based on a user's interaction with an item (e.g. movie watched or not, product bought or not, watch time, etc) can have many confounding factors, and it is difficult to get feedback about regretted interactions this way. Numerical explicit, e.g. a 5-points scale enables systems to ignore the non-rated items - but also provides a sparser signal. Explicit feedback is not limited to quantitative assessment. The Web has facilitated immediate exchanges of opinion on various topics in online fora; social networks\footnote{\url{https://www.facebook.com/}, \url{https://twitter.com/}}, e-commerce\footnote{\url{https://www.amazon.com/}, \url{https://www.ebay.com/}} and streaming service\footnote{\url{https://www.netflix.com/}} platforms democratized it, and it is now common for users to provide feedback online for the services and products they consumed. 


Contrary to tags or movie plot descriptions, text reviews are totally free-form and may contain many levels of preference information. Text reviews include sentences useful to represent users' tastes or movie features as shown in Table~\ref{tab:first_examples}. Some may even contain a direct explicit recommendation from a human, which is likely very valuable information for rating prediction.
This motivates using free-form text reviews in the design and explanation of rating prediction systems.

\begin{table}
  \centering
  \footnotesize
  \begin{tabular}{p{2.5cm} p{4cm}} \toprule
    \multicolumn{1}{p{2.5cm}}{Item \itemj liked by some user \useru} &  \multicolumn{1}{p{4cm}}{Other item reviewed by \useru, along with a sentence \useru wrote} \\\midrule

    \multirow{3}{*}{\parbox{2cm}{\raggedright Sleeping beauty}}  & \textbf{Ratatouille} --  `` Ratatouille '' is a thoroughly entertaining movie that is perfect family fare for the summer . \\
    
    \multirow{2}{*}{\parbox{2cm}{\raggedright The Shining}}  & \textbf{Haloween} --  This is a must movie for all true horror fans . \\
    
    \multirow{4}{*}{\parbox{2cm}{\raggedright Groundhog day}}  & \textbf{The Truman Show} --  `` The Truman Show '' will make you laugh , and keep you on the edge of your seat , wondering if Truman will ever get out. \\\bottomrule

  \end{tabular}
  \caption{Examples of sentences (right column) written by a user \useru on some item (in bold) they reviewed and useful to infer that \useru likes another item \itemj (left column). An item is ``liked'' when the rating is above $4/5$.}
  \label{tab:first_examples}
\end{table}

\section{Datasets}
We experimented with the 2014 Amazon Product Review dataset \citep{10.1145/2872427.2883037}, made of reviews from users on items with 5-points scale ratings and text reviews. It has been used in related work \citep{10.1145/3331184.3331267, 10.1145/3292500.3330989, 10.1145/3219819.3219823} for its important size. Specifically, we focused on the movie subset because common knowledge makes it easier to interpret movie reviews than reviews of other items. Following \citet{he2017neural, DBLP:journals/corr/abs-2109-09358} we restricted our study on the k-core subset, where $k=20$. Additionally, we trained and evaluated our method on the k-core subset of the Yelp review dataset \cite{yelp}, made of reviews of businesses (like restaurants). Table~\ref{tab:dataset_stats} shows statistics for the datasets.


\begin{table}
  \centering
  \small
  \begin{tabular}{cS[table-format=6.1]S[table-format=7.1]}
  \toprule
    \textbf{Dataset} & \textbf{Amazon 20-core} &  \textbf{Yelp 20-core} \\ \midrule
    Users & 3728 & 38595 \\
    Items & 3911 & 27823 \\
    Train reviews & 205158 & 1929332 \\
    Validation reviews & 3728 & 38595 \\
    Test reviews & 3728 & 38595  \\
    Average sentence \# per review & 14.1 & 9.8 \\
    Average token \# per sentence & 46.2 & 30.5 \\\bottomrule 
  \end{tabular}
  \caption{Statistics for the 2014 Amazon Movie Review dataset and the Yelp Review dataset. Following \citet{yang-etal-2020-multilingual}, sentences are tokenized with SentencePiece \citep{kudo-richardson-2018-sentencepiece}.}
  \label{tab:dataset_stats}
\end{table}

The datasets are split into training, validation, and test subsets. Various splitting strategies exist \citep{10.1145/3383313.3418479}, showing benefits and limitations depending on the nature of the feedback (explicit or implicit), of the task (rating prediction or item recommendation) and the model used. Following previous works \citep{3195c13d3d90462f8aa02b15d6e7d35d, DBLP:journals/corr/abs-2109-09358}, we preserve time ordering \citep{Shani2011} by employing the \textit{Leave One Last Item} strategy. The test and validation sets are respectively made of all users' last and penultimate interactions; the training set consists of the remaining interactions. In the next section, we describe how a Universal Sentence Encoder (USE) can be used to build representations of sentences found in reviews.   

\section{Semantic matches} \label{sec:semantic_match}
\subsection{Semantic representations of reviews} \label{sec:semantic_representation}
The Universal Sentence Encoder (USE) is a sentence embedding model pre-trained on a variety of NLP tasks with diverse degrees of supervision, namely Retrieval Question-Answering, Translation Ranking, and Natural Language Inference \citep{bowman-etal-2015-large}. It can be thought of as a paraphrase similarity model and \citet{yang-etal-2020-multilingual} showed such embeddings were useful for a variety of downstream NLP tasks. We encoded review sentences in a single high-dimensional space using the USE model. 
To illustrate the intuition for the relevancy of USE on review sentences, we applied it to three popular and distinct categories of items. We selected three items per category, according to the genre tag found on the Internet Movie Database\footnote{\url{https://www.imdb.com/}}. The sentences found in the train reviews of the nine items are embedded in the USE model's $512$ dimensional semantic space and we show in Figure~\ref{fig:tsne} their 2D t-SNE \citep{JMLR:v9:vandermaaten08a} projection. We note that individual items have clustered sentences. On top of that, we observe that the projections of categories are Y-shaped indicating some clustering at the category level as well, though sentences at the origin seem to share a high similarity across categories. We note many of the latter sentences, with low semantic variability, are general sentences, not specific to any item, such as ``I liked it''.


\subsection{Aggregation of matches per-item category} \label{sec:category_matches}
\begin{figure*}
    \centering
    \includegraphics[width=0.73\textwidth]{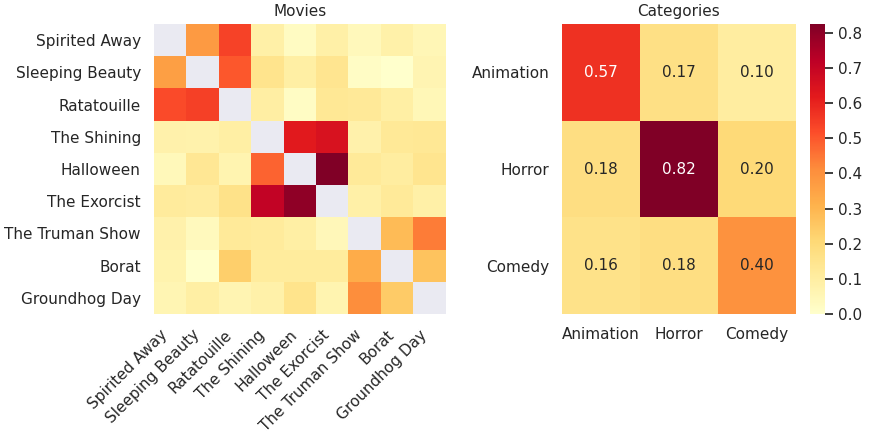}
    \caption{Heatmaps of k-NN sentence embedding matches where $k=10$. Rows and columns respectively represent head and tail vertices. Same-item matches are discarded. 
    Left: matches aggregated per movie. Right: matches aggregated per category.}
    \Description{Fully described in the text.}
    \label{fig:heatmaps}
\end{figure*}

We computed the k-NN sentence graph \globalgraph of all the sentences found in the 20-core Amazon movie review trainset, embedded with USE. It is directed since each edge leaves sentence \sentenceu and enters their k nearest neighbors \sentencev. Following the approach of \citet{yang-etal-2020-multilingual} for transfer learning, we used the cosine distance (defined as $1$ minus the cosine similarity)\footnote{USE's embedding are normalized on the unit sphere.}. We call ``semantic match'' a pair of connected sentences in the graph.  
Figure~\ref{fig:heatmaps} shows the heatmaps of matches in the sub-graph of the nine mentioned items, aggregated by items and categories. Since the items have different numbers of train sentences, the match count matrix is normalized with the square root of its row and column weights \citep{hirschfeld_1935}.
It appears that the relative number of matches is higher for pairs of sentences belonging to the same category of items.

Visualization and quantitative analysis of matches indicate that one can represent item reviews with USE and build a rating predictor from it.

\section{Text-based k-NN rating prediction systems} \label{sec:rec_sys}
Text-KNN estimates the test rating as the following weighted sum:
\begin{equation} \label{eq:rec_sys}
\hat{r}_{ui} = \frac{
        \sum\limits_{v \in N^{k'}_i\left(u\right)} \usersim \cdot r_{vi}}
        {\sum\limits_{v \in N^{k'}_i\left(u\right)} \usersim}
\end{equation}
where \usersim is the weight of \userv for predicting \useru's rating. Among all the possibilities to compute \usersim, we present below three alternatives based on counting the semantic matches. $N^{k'}_i$ is the set of $k'$ nearest neighbor users of user $u$ who have rated item $i$.

A simple method, called ``baseline'' in our experimental setups and introduced by \citet{10.1145/1644873.1644874}, estimates ratings $b_{ui}$ as the overall average $\mu$ rating summed to user and item biases (respectively $b_u$ and $b_i$). We experimented with the baseline-aware user k-NN (Text-BKNN) variant: 
\begin{equation} \label{eq:rec_sys_b}
\hat{r}_{ui} = b_{ui} + \frac{ \sum\limits_{v \in N^{k'}_i\left(u\right)}
        \usersim \cdot \left(r_{vi} - b_{vi}\right)} {\sum\limits_{v \in
        N^{k'}_i\left(u\right)} \usersim}
\end{equation}

We propose a user similarity measure based on the representations of sentences seen in Section~\ref{sec:semantic_representation}. 

\paragraph{Notations} 
Let $\sentencegraph=\left(V,E\right)$ be the k-NN directed graph of sentences. Each directed edge $\left(\sentenceu, \sentencev\right) \in \edges$ has an associated semantic similarity weight $\edgeweight \in \left[0,1\right]$, detailed in Section~\ref{sec:s}. By extension, $\edgeweight = 0$ if $\left(\sentenceu, \sentencev\right) \notin \edges$.
The three approaches experimented to compute \usersim, denoted ``One-to-One'', ``Many-to-One'' and ``Many-to-Many'', rely on the fact that the set of sentences is partitioned by the set of users. 
For that reason, let \usentences be the set of sentences written by \useru.\\
$\userneighborhoodsentence = \left\{ \userv \in \users : \exists \sentencev \in S_{\nu}, \edgeweight > 0 \right\}$ is the set of users \userv whose sentences appear at least once in the neighborhood of sentence \sentenceu. \\
$\userneighborhooduser = \bigcup_{\sentenceu \in \usentences} \userneighborhoodsentence$ is the set of users \userv whose sentences appear at least once in the neighborhood of sentences written by user \useru. Indicators are denoted by \indicator.

\paragraph{One-to-One matching}
Here, we consider \useru and \userv to have expressed similar preferences if \userv wrote at least one sentence in the semantic neighborhood of at least one sentence written by \useru, i.e.
\begin{equation} \label{eq:1_1}
    \graphusersimoneone = \indicator\left\{\userv \in \userneighborhooduser\right\}
\end{equation}

\paragraph{Many-to-One matching}
Alternatively, we computed the occurrence of \userv's sentences in the neighborhood of \useru's sentences, i.e.
\begin{equation} \label{eq:n_1}
    \graphusersimmanyone = \sum\limits_{\sentenceu \in \usentences} \indicator\left\{ v \in \userneighborhoodsentence \right\}
\end{equation}

\paragraph{Many-to-Many matching}
Our third approach consisted in counting the number of sentence matches when considering all pairs of sentences written by \useru and \userv. In this case,

\begin{equation} \label{eq:n_n}
    \graphusersimmanymany = \text{max}\left[ \sum\limits_{\left(\sentenceu, \sentencev\right) \in \left(\usentences \times \vsentences \right)} \edgeweight, 0\right]  
\end{equation}

As we shall see in the next section, \edgeweight is not necessary non-negative, though Equations~\eqref{eq:rec_sys} and~\eqref{eq:rec_sys_b} require non-negative weights.

\section{Sentence graph} \label{sec:s}
The weight \edgeweight introduced in section \ref{sec:rec_sys} corresponds to the semantic similarity of sentence \sentencev regarding \sentenceu. Again, we considered various ways of computing it: ``binary'', ``continous'' and ``polarized''.

\paragraph{Binary count}
The basic sentence weight consists in counting $1$ if and only if the edge from \sentenceu to \sentencev appears in the graph.




\paragraph{Continuous similarity scores}
To include finer information about sentence similarity, we experimented with the edge weight to be the cosine similarity scaled between $0$ and $1$. We defined $s$ by: 
\begin{align*}
\edgeweight = \frac{1+\cosinesim\left(\sentenceu, \sentencev\right)}{2}
\end{align*}



\paragraph{Polarization}
Supplementary information at the sentence level may be relevant where semantic matching fails. For instance, the sentences ``I love DiCaprio'' and ``I hate DiCaprio'' have a cosine similarity of $0.94$. We propose to integrate information from a sentiment attribution mechanism $\sentiment(\sigma) \in \left[-1,1\right]$ with a polarization function $p$ defined in equation~\eqref{eq:polarization}. Then, the sentence weight becomes $p\cdot s$.

\begin{equation} \label{eq:polarization}
\polarization =
\begin{cases}
    \hphantom{-}1,  & \text{if } |\sentiment (\sentenceu) - \sentiment(\sentencev)| \leq 1 \\
    -1, & \text{otherwise}
\end{cases}
\end{equation}

We experimented with a rating-aware sentiment attribution defined in equation~\eqref{eq:sentiment}.
\begin{equation} \label{eq:sentiment}
\text{\sentiment}(\sigma)= 
    \begin{cases}
        \hphantom{-}1,  & \text{if } r\left(\sigma\right) \geq 4 \\
        -1, & \text{if } r\left(\sigma\right) \leq 2 \\
        \hphantom{-}0,  & \text{otherwise}
    \end{cases}
\end{equation}
$r(\sigma)$ being the rating of the review $\sigma$ belongs to.

\paragraph{Item graphs}
Moreover, we experimented with the methods discussed above considering graphs made of sentences of a single item, rather than a single ``global'' graph of sentences from all items.
The final user weight \usersim is the sum of the per-item user weights, though other aggregations may be considered.

\paragraph{Normalization} \label{sec:normalization}
Some users write more sentences than others, some items receive more sentences than others and users may write different numbers of sentences on different items. For all these reasons, bias may appear when the user weights are computed from counting semantic matches between sentences (or co-occurrence like in the ablation study in Section~\ref{sec:discussion}). To mitigate this potential source of bias, we integrated different ad hoc normalizations in the computation of the user weights. 

On the one hand, we tried to normalize \usersim by the number of sentences $|\vsentences|$ written by \userv. In addition to this option, we considered normalizing the weights by the number of sentences $|\isentences|$ written on \itemi or the number of sentences $|\uisentences|$  written by \useru on \itemi  when per-item graphs were considered. On the other hand, bias induced by an imbalance of sentence set sizes could also be mitigated by normalizing the match counting methods introduced in section \ref{sec:rec_sys}. Concerning the One-to-One matching (Equation~\eqref{eq:1_1}), we tried to normalize by \userneighborhooduser. When considering Many-to-One matching, each term in Equation~\eqref{eq:n_1} was divided by $1$ or \userneighborhoodsentence. For the Many-to-Many matching, the terms in the Equation's~\eqref{eq:n_n} sum are divided either by $1$, $\weightedindegree = \sum_{\sigma \in \vertices} |s\left(\sigma, \sentencev\right)|$ or $\weightedoutdegree=\sum_{\sigma \in \vertices} |s\left(\sentenceu, \sigma\right)|$. The latter two options respectively correspond to extensions of in and out-degrees adapted to our weighted graph. Normalizing by the degrees is unsupervised mitigation of semantic similarity between ``common'' sentences, which are irrelevant for the sake of representing user preferences (cf. Section~\ref{sec:semantic_representation}).

\section{Evaluation measures} \label{sec:evaluation}
Offline evaluation of rating prediction is popularly measured through RMSE. However, \citet{10.1145/1864708.1864721} showed the limits of pure error-based metrics. 

As an example, consider the toy dataset in Table~\ref{tab:metrics_example}. Neighborhood-based CF clearly indicates that $u_2$ behaves like $u_0$ and the ground truth tells that $u_2$ prefers $i_1$ over $i_0$.

The baseline asymptotically predicts $\lim_{n\to\infty}\hat{r}_{u_2i_0}(n)=3^+$ and  $\lim_{n\to\infty}\hat{r}_{u_2i_1}(n)=3^-$.
Therefore, it yields $\lim_{n\to\infty}\text{RMSE}(n)=0.5$, relatively good compared to the uniform random ($\mathbb{E}\left[\text{RMSE}\right]=\frac{17}{16}$). However, the baseline is non-personalized when it comes to compare rankings, and always predicts $\hat{r}_{u_2i_0}(n) > \hat{r}_{u_2i_1}(n)$. This example illustrates RMSE's limitation in reflecting user preferences modeled by systems. This motivates a ranking-based metric measuring how well the ordering of items is preserved by the system.

\begin{table}
  \centering
  \small
  \begin{tabular}{c|cccc}
  \toprule
    \backslashbox{User}{Item} & $i_0$ & $i_1$ & $i_2$ & $i_3$\\\midrule
    $u_0$ & 2 & 4 & 1 & 5 \\
    $u_1$ & 5 & 1 & 5 & 1 \\
    $u_2$ & \textbf{2.5} & \textbf{3.5} & 1 & 5 \\
    $u_3 \hdots u_n$ & 5 & 5 & 5 & 5 \\\bottomrule
  \end{tabular}
  \caption{Toy dataset of ratings. Bold values are in the test set and the remaining values are in the train set.}
  \label{tab:metrics_example}
\end{table}

We argue that the Fraction of Concordant Pairs (FCP), measuring the proportion of well-ranked item pairs, is a suitable metric for three reasons. First, it directly and unambiguously measures preferences expressed by users, by definition. Second, it is grounded in statistics since concordant and discordant pairs have already proved relevant to compare two measured quantities \citep{10.1093/biomet/30.1-2.81}. Third, it generalizes to non-binary ordered sets the ROC-AUC binary metric, known to test whether positive examples are ranked higher than negative examples by classifiers.

We handle equalities the same way as \citet{Surprise}, i.e.\\ $\left(\left(r_{ui}, \hat{r}_{ui}\right), \left(r_{uj}, \hat{r}_{uj}\right)\right)$ is:
\begin{itemize}
    \item Concordant (CP) iif $r_{ui} \neq r_{uj}$ and\\ $\text{sgn}\left(r_{ui} - r_{uj}\right) = \text{sgn}\left(\hat{r}_{ui} - \hat{r}_{uj}\right)$,
    \item Discordant (DP) iif $r_{ui} \neq r_{uj}$ and\\ $\text{sgn}\left(r_{ui} - r_{uj}\right) \neq \text{sgn}\left(\hat{r}_{ui} - \hat{r}_{uj}\right)$,
    \item Ignored if $r_{ui} = r_{uj}$
\end{itemize}

In the toy example, the baseline has a FCP of 0, signaling its inability to model $u_2$'s preferences. 

\citet{Koren2013CollaborativeFO} randomly split the dataset into train and test data, enabling the computation of FCP on pairs of test items rated by the same user. Yet, modern time-based \textit{Leave One Last Item} splitting strategies provide one single test item per user. We adapted FCP to consider, for each user, all the pairs made of the test item and a train item. We call it \textit{Time-based FCP} (\lfcp). Thereby, the evaluation is equivalent to assessing the frequency of the model to correctly rank the next rating compared to all past ratings, for each user. Denoting respectively $n_c$ and $n_d$ the number of concordant and discordant pairs, we report the per-user macro-averaged metric: $\lfcpmacro=\sum\limits_{\useru\in\users}\frac{n_c\left(u\right)}{n_c\left(u\right)+n_d\left(u\right)}$. 

\section{Results}
\subsection{Experimental setup}
\begin{figure*}[h]
    \centering
    
    \begin{minipage}{.3\textwidth}
      \centering
      \ \ \ \ \ \ \ \ \ \ \ \ \ \ \ \ Amazon \\
      \includegraphics[width=1.0\linewidth]{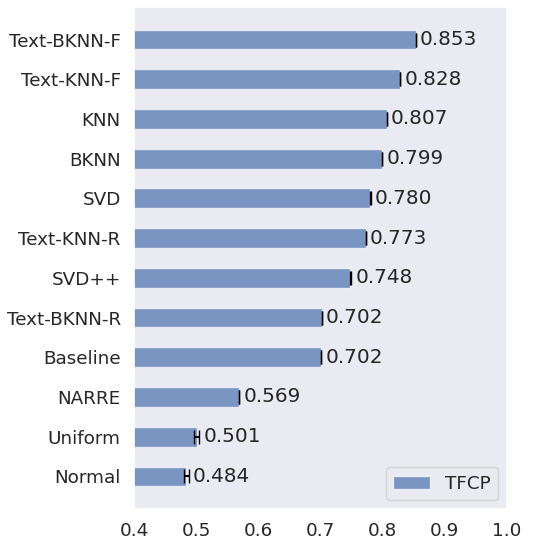}
    \end{minipage}%
    \begin{minipage}{.3\textwidth}
      \centering
      \ \ \ \ \ \ \ \ \ \ \ \ \ \ \ \ \ \ Amazon \\
      \includegraphics[width=1.0\linewidth]{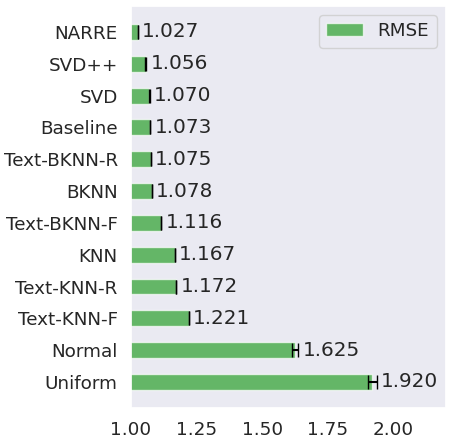}
    \end{minipage}
    \begin{minipage}{.3\textwidth}
      \centering
      \ \ \ \ \ \ \ \ \ \ \ \ \ \ \ \ \ \ Yelp \\
      \includegraphics[width=1.1\linewidth]{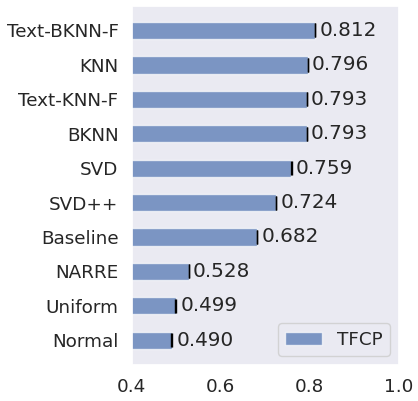}
    \end{minipage}
    \Description{Fully described in the text.}
    \caption{Bar charts of the models' test scores (average and standard deviation) on the Amazon and Yelp datasets.}
    \label{fig:results20core}
\end{figure*}

We ran a grid search (corresponding to 276 search trials) over the set of parameters and options described in the previous section. For both Text-KNN and Text-BKNN, we selected the models yielding the best validation RMSE (-R), and \lfcpmacro (-F) on the Amazon 20-core dataset. The set of parameters resulting from the tuning is found in Table~\ref{tab:hyperparameters} (cf. the appendix) and discussed in Section~\ref{sec:discussion}. We compared our approach to three baselines (two random systems: Uniform, Normal, and a popularity-based baseline, defined in Section~\ref{sec:rec_sys}), two popular memory-based methods (the rating-based KNN and BKNN, with MSD as similarity) and three state-of-the-art\footnote{Relative to error-based metrics and rating prediction tasks.} model-based systems (SVD, SVD++ and NARRE \citep{10.1145/3178876.3186070}). NARRE is a deep learning model relying on a neural attention mechanism to concurrently make recommendations and decide which reviews are most relevant. Unless specified otherwise, their hyperparameters are the default parameters from \citet{Surprise}. Specifically, all k-NN regressors had a number of \textit{user} neighbors $k'=40$ and text k-NN graphs used a number of \textit{sentence} neighbors $k=10$. For random baselines, SVD and SVD++, we repeated 10 times training and evaluation with different random seeds. In order to assess how the best set of hyperparameters found on the Amazon dataset generalizes to other datasets, we trained and evaluated Text-KNN-F and Text-BKNN-F on the Yelp dataset, i.e. without any hyperparameter tuning specific to that dataset. Figure~\ref{fig:results20core} compares the performances of different models over both metrics.


\subsection{Discussion} \label{sec:discussion}
\begin{figure}
    \includegraphics[width=0.49\textwidth]{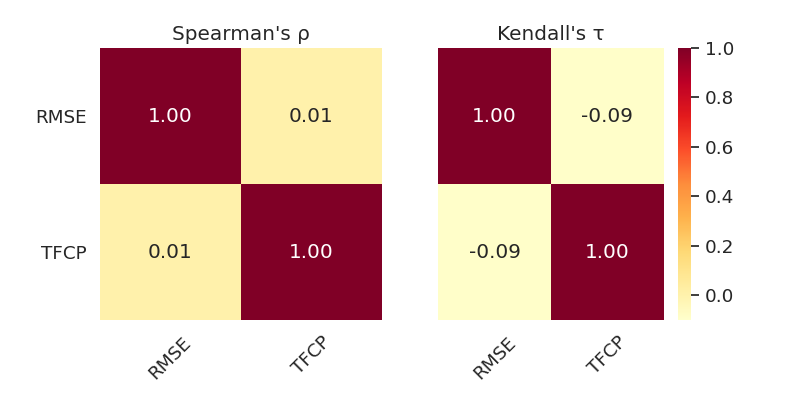}
    \caption{Correlation coefficients of the rankings produced by the metrics when considering the systems in Figure~\ref{fig:results20core}.}
    \Description{Fully described in the text.}
    \label{fig:correlation}
\end{figure}

Quantitative evaluation indicates that our approach is comparable to previous text-agnostic memory-based systems for both RMSE and TFCP. A general trend is that model-based systems give better RMSE but worse TFCP. Surprisingly, RMSE suggests the baseline's performance is better than all memory-based systems, although being a naive approach. This echoes our analysis in Section~\ref{sec:evaluation} and validates the motivation behind the ranking-based evaluation. TFCP does indeed rank the non-``rankingwise personalized'' baseline in penultimate place among non-random systems. Similarly, NARRE outperformed all other systems when evaluated with RMSE while being only slightly better than random predictions regarding TFCP. For further analysis of the correlation between RMSE and TFCP, Figure~\ref{fig:correlation} shows Spearman's $\rho$ and Kendall's $\tau$ correlation coefficients of the rankings produced by the metrics. RMSE and TFCP produce decorrelated rankings.

After hyperparameter tuning on the Amazon dataset, our systems are ranked first on TFCP, while showing competitive RMSE scores. In particular, the metric ranks text-based systems as the best among memory-based. Even when trained and evaluated on the Yelp dataset our models rank first and third if we do not tune the hyperparameters specifically for that dataset.

Furthermore, results show that one can tune our approach's hyperparameters to optimize each metric. Doing so enables text-based models to yield better evaluation scores than their text-agnostic versions (e.g. once optimized for a metric, text-BKNN has a better score than BKNN), except for KNN on the RMSE metric. There is no winner-take-all hyperparameter set, achieving the best results on both RMSE and TFCP. We observed that three out of four of the tuned text-based models involved the Many-to-Many matching and polarization. The latter signals the relevancy of sentiment-awareness for sentence comparison. Incidentally, the TFCP-optimized text-based models with rating-agnostic computation of user similarity (i.e. without polarization) had $\lfcpmacro = 0.849$ when trained and evaluated on the Amazon dataset. When optimizing \lfcpmacro, both Text-KNN and Text-BKNN make the same set of hyperparameters stand out (cf. Table~\ref{tab:hyperparameters}'s second row in the appendix). According to this metric, the best way of computing user similarity with semantic matches involves the fine-grained Many-to-Many match count, with continuous similarity and polarization information, when per-item graphs are considered. On the contrary, the best Text-based model for RMSE (Text-BKNN-R) is obtained with the coarse-grained One-to-One matching, minimal information from the binary count. It is worth noting that in the latter case, the global graph is preferred. Plus, we also considered filtering out matches with sentence similarity below a certain threshold, but interestingly, early experiments showed no quantitative benefit.




\paragraph{Ablation study} We observed that our best text-based models involved per-item graphs instead of the global graph. To assess the benefit of semantic similarity, we studied the performance of three naive implementations counting the co-occurrence of sentences written by a pair of users \useru and \userv on an item \itemi. The weights \usersim was computed by summing user weights $w_i\left(\useru,\userv\right)$ over all items \itemi. The implementations we tried were:

\begin{itemize}
    \item $w_i\left(\useru,\userv\right)=\indicator\left\{|\uisentences|>0 \text{ and } |\visentences|>0 \right\}$.

    \item $w_i\left(\useru,\userv\right)=|\uisentences|\cdot\indicator \left\{|\visentences|>0 \right\}$.

    \item $w_i\left(\useru,\userv\right)=|\uisentences|\cdot|\visentences|$.
\end{itemize}

After selecting the best option on the Amazon validation set, we found the best test scores to be $\text{RMSE} = 1.197$ and $\lfcpmacro = 0.787$. Even though the evaluation scores indicate existing signal from this similarity-agnostic approach, we see significant improvement and interest in integrating semantic similarity and finer relationships between sentences.

\paragraph{Explainability}
Besides quantitative performances, our text-based approach has benefits in terms of explainability of automatic recommendation. Table~\ref{tab:explainability_global_different}, Table~\ref{tab:explainability_global_same_short} and Table~\ref{tab:explainability} (cf. the appendix) show pairs of sentences from the Amazon train reviews written by some user \useru and \useru's nearest user neighbor \userv. Both users \useru and \userv liked \useru's test item \itemj, i.e. $r_{uj} \geq 4$ and $r_{vj} \geq 4$. The authors manually selected sentence pairs counted as semantic matches by the system, and relevant for explaining why the system predicts that \useru's opinion on \itemj ($\hat{r}_{uj}$) should resemble \userv's opinion on \itemj ($r_{vj}$), through a high user similarity \usersim. 
For readers unaware of the items, the authors added relevant attributes and commonalities relative to items. Table~\ref{tab:explainability} corresponds to the case of per-item graphs while table~\ref{tab:explainability_global_different} and table~\ref{tab:explainability_global_same_short} gather matches between sentences from different items in the global graph. The latter tables differ from each other in whether tail sentences (written by \userv) review \useru's test item \itemj or not, which is equivalent for the rating prediction system but may matter in human interpretation.

\begin{table*}
  \small
  \begin{tabular}{p{2cm} >{\raggedright}p{2.3cm} p{12.2cm}} \toprule
    \multicolumn{1}{p{2cm}}{\useru's test item \itemj} & Attribute of \itemj &  \multicolumn{1}{c}{Matching sentences in the train set} \\\midrule

    \multirow{2.5}{*}{\parbox{2cm}{\raggedright Hancock}} &  \multirow{2.5}{*}{\parbox{2.3cm}{\raggedright Action film}} & \textbf{To Live} -- Usually I ' m bouncing off the walls and watching a crazy action flick .\\ \cmidrule[0.0pt](l){3-3}
    & & \textbf{Gladiator} -- Sometimes I just fast forward straight to the epic battle scenes .\\\midrule

    \multirow{2.5}{*}{\parbox{2cm}{\raggedright Non-Stop}} &  \multirow{2.5}{*}{\parbox{2.3cm}{\raggedright Action thriller film}} & \textbf{The Debt} -- Plot is very slow developing , so slow and uninteresting , did not even finish watching .\\ \cmidrule[0.0pt](l){3-3}
    & & \textbf{Whiteout} -- Plot was so S-L-O-W and dull .\\\midrule

    \multirow{2.5}{*}{\parbox{2cm}{\raggedright Dune}} &  \multirow{2.5}{*}{\parbox{2.3cm}{\raggedright Action-adventure science-ficton film}} & \textbf{Oblivion} --  Good action and story line .\\ \cmidrule[0.0pt](l){3-3}
    & & \textbf{Serenity} -- great action and story line .\\\midrule

    \multirow{2.5}{*}{\parbox{2cm}{\raggedright Evil Dead}} &  \multirow{2.5}{*}{\parbox{2.3cm}{\raggedright Horror film}} & \textbf{Suspiria} --  A legend in the horror movie genre .\\ \cmidrule[0.0pt](l){3-3}
    & & \textbf{Prince of Darkness} -- A horror movie classic .\\\midrule

    \multirow{3.5}{*}{\parbox{2cm}{\raggedright Rocky II}} &  \multirow{3.5}{*}{\parbox{2.3cm}{\raggedright Written by and starring Sylvester Stallone}} & \textbf{Rambo III} --  His range may be limited , as we saw in his ' comedy ' films , but when he sticks to his forte , playing great heroes , Stallone is the greatest ever .\\ \cmidrule[0.0pt](l){3-3}
    & & \textbf{Rocky} -- But Stallone gives just about the best performance of his career here .\\\midrule

    \multirow{4.5}{*}{\parbox{2cm}{\raggedright The Living Daylights}} &  \multirow{4.5}{*}{\parbox{2.3cm}{\raggedright Entry in the James Bond (a.k.a. \textit{007}) series}} & \textbf{Never Say Never Again} --  " Never Again " ultimately retains a very watchable magic featuring the original Agent 007 one last time .\\ \cmidrule[0.0pt](l){3-3}
    & & \textbf{Goldfinger} -- After the first two 007 films , this third Bond adventure cemented forever the style and fun of the series .\\\midrule

    \multirow{2.5}{*}{\parbox{2cm}{\raggedright My Neighbor Totoro}} &  \multirow{2.5}{*}{\parbox{2.3cm}{\raggedright Animated film}} & \textbf{Popeye} --  Hey parents , want a good , clean , wholesome movie for your kids ?\\ \cmidrule[0.0pt](l){3-3}
    & & \textbf{Kiki's Delivery Service} -- An excellent movie for kids that parents don ' t have to worry about .\\\midrule
    
  \end{tabular}
  \caption{Examples of semantic matches when our system considers the \textbf{global graph}. The last column shows the head sentence first (written by \useru) and the tail sentence then (written by \userv), along with the respective item's review they belong to. Here, tail sentences do not review \itemj but express attributes also present in \itemj and indicated in the second column.}
  \label{tab:explainability_global_different}
\end{table*}

\begin{table*}
  \centering
  \small
  \begin{tabular}{p{9cm} p{8cm}} \toprule
    \textsc{Head sentence \sentenceu written by \useru on some item \itemi} & \textsc{Tail sentence \sentencev written by \userv on \useru's test item \itemj and matching \sentenceu} \\\midrule
    
    \textbf{Cemetery Man} -- This highly entertaining little zombie movie from Italy has all the elements that make it a wonderfully dark horror - comedy in the same vein asEvil Dead 2 : Dead by DawnandAn American Werewolf in London . 
    & \textbf{The Horde} -- one of the best horror zombie movies of all the times , this movie is equal than 28 days later , in these days European horror movies are the best of the best . . good for Friday at night \\\hline
    
    \textbf{Charlie's Angels} -- I like drew barrymore , she is the best angel out of the three.
    & \textbf{Fever Pitch} -- Drew Barrymore , as always , is phenomenal . \\\hline

    \textbf{Ice Age} --  While not perfect , it is full of laughs and beautiful computer animation.
    & \textbf{Finding Nemo} -- Highlights : Spectacular computer animation ; hilarious , well - developed characters ; original plot .\\\hline

    \textbf{Monsters, Inc.} -- Great for parents and kids ( or people without kids ) .
    & \textbf{Toy Story 3} -- Great for kids and adults alike .\\\hline

    \textbf{Island in the Sky} -- The Duke\footnotemark does a great job in his role as Dooley - the plane's captain .
    & \textbf{The Alamo} -- The Duke turns out one of his best performances , as well as putting together this film .\\\hline

     \textbf{I Am Legend} -- At the end of the movie you fail to realise it was just one man , Will Smith in most of the scences and yet the movie is neither boring , or lacking in elements that make for a great thriller .
    & \textbf{Hancock} -- Will Smith adds a lot of flare to the movie , when it could ' ve been bland and cheesy .\\\hline
    
    \textbf{War of the Worlds} --  The special effects were great and the acting was believable.
    & \textbf{Munich} -- The acting , story line and special effects were great .\\\midrule
    
    \textbf{Red River} --  I think this is my favorite early John Wayne film ( it's not exactly early , but it was one of his earlier big hits ) .
    & \textbf{The Searchers} -- This is John Wayne's favorite " John Wayne movie , " and his acting is superb .\\\hline
    
    \textbf{Bubba Ho-Tep} -- Fans of Bruce Campbell will love this movie , but I don ' t know how fans of the King will take it .
    & \textbf{My Name Is Bruce} -- First of all if your not a bruce campbell fan or dont enjoy his movies then you probably wont like this one as it is trademark campbell . . . .\\\bottomrule
  \end{tabular}
  \caption{Examples of semantic matches when our system considers the global graph. Sentences are preceded by the item they review. \userv represents \useru's nearest user neighbor. Here, tail sentences review \itemj but tail sentences reviewing other items can count as semantic matches.}
  \label{tab:explainability_global_same_short}
\end{table*}
\footnotetext[8]{A nickname for the American actor John Wayne}

\section{Conclusions \& Further work}

This study explored, for the first time, the application of pre-trained NLP models of semantic similarities to the task of rating prediction. The experiments quantitatively show benefits of using review text in memory-based models. Furthermore, the representation of user-preferences as textual snippets provides a concrete implementation for better explainability of rating prediction systems. Finally, this study also highlighted the robustness of an FCP-based ranking-based metric for rating-prediction, inspired by the observation that different users use rating scales in different ways.

Future work includes evaluating explanations from our approach as well as building better representations of users and items. For example, language models could be jointly tuned with a recommendation task (cf. Section~\ref{domain_specific} in the appendix), unsupervised techniques could help filtering out irrelevant or redundant text, and rationale extraction like that of WT5 \citep{narang2020wt5} could preprocess sentences to make better and more concise textual preference representations. More generally, our study highlights an exciting and emerging intersection between the recent advances in natural language processing and rating prediction systems.

\bibliographystyle{ACM-Reference-Format}
\bibliography{www}

\appendix

\begin{table*}[h!]
  \centering
  \small
  \begin{tabular}{p{3cm} c >{\raggedright}p{2.3cm} p{9cm}} \toprule
    \multicolumn{1}{p{2cm}}{\useru's test item \itemj} & \multicolumn{1}{p{2cm}}{Train item \itemi co-reviewed by \useru and \userv} & Commonality shared by both \itemi and \itemj &  \multicolumn{1}{c}{Matching sentences in the train set} \\\midrule

    \multirow{2.5}{*}{\parbox{3cm}{\raggedright Justice League: The Flashpoint Paradox}} & \multirow{2.5}{*}{\parbox{2cm}{\raggedright Green Lantern: First Flight}} & \multirow{2.5}{*}{\parbox{2.3cm}{\raggedright Animated superhero film}} & You simply must check out this awesome superhero epic .\\ \cmidrule[0.0pt](l){4-4}
    & & & Personally , it's the finest superhero animated effort ever brought to screen .  \\\midrule

    \multirow{3.5}{*}{\parbox{2cm}{\raggedright  Fort Apache}} & \multirow{3.5}{*}{\parbox{2cm}{\raggedright The Searchers}} & \multirow{3.5}{*}{\parbox{2.3cm}{\raggedright Western film directed by John Ford and starring John Wayne}} & That aside , and at the risk of repeating myself it's vintage John Ford and John Wayne with some magnificent scenery .\\ \cmidrule[0.0pt](l){4-4}
    & & & This is one of John Ford's and one of John Wayne's best movies . \\\noalign{\vskip 1mm}\midrule

    \multirow{3.5}{*}{\parbox{2cm}{\raggedright Peter Pan}} & \multirow{3.5}{*}{\parbox{2cm}{\raggedright Monsters, Inc.}} & \multirow{3.5}{*}{\parbox{2.3cm}{\raggedright Animated film}} & This is a movie that I ' d suggest any family to pop in and have a family movie night .\\ \cmidrule[0.0pt](l){4-4}
    & & & a great family movie .\\\midrule
    
    \multirow{2.5}{*}{\parbox{2cm}{\raggedright Killing Them Softly}} & \multirow{2.5}{*}{\parbox{2cm}{\raggedright Django Unchained}} & \multirow{2.5}{*}{\parbox{2.3cm}{\raggedright Violence \& Gore}} & This movie is Bloody , gory , violent , emotional , serious and hilarious .\\ \cmidrule[0.0pt](l){4-4}
    & & & All in all . . . . a really good movie , but a bit bloody .\\\midrule

    \multirow{2.5}{*}{\parbox{2cm}{\raggedright 1776}} & \multirow{2.5}{*}{\parbox{2cm}{\raggedright Downfall}} & \multirow{2.5}{*}{\parbox{2.3cm}{\raggedright Historical drama film}} & It is almost a must see for historians and those with an interest in history .\\ \cmidrule[0.0pt](l){4-4}
    & & & Should be required viewing for high school history students .\\\bottomrule

  \end{tabular}
  \caption{Examples of semantic matches when our system considers \textbf{per-item graphs}. The last column shows the head sentence first (written by \useru) and the tail sentence then (written by \userv).}
  \label{tab:explainability}
\end{table*}
\FloatBarrier
\section{Appendix}
\begin{table*}
  \centering
  \small
  \begin{tabular}{l|cccccc|}
  \toprule
    \backslashbox{Model}{Option} & User similarity & Sentence weight  & Polarization & Normalization & Type of graph\\\midrule
    Text-KNN-R   & Many-to-Many & Continuous  & Yes & Out-degree & \multirow{2}{*}{Global graph} \\
    Text-BKNN-R  & One-to-One & Binary &  No & 1 & \\\hline
    Text-KNN-F  & \multirow{2}{*}{Many-to-Many} & \multirow{2}{*}{Continuous}  & \multirow{2}{*}{Yes} & \multirow{2}{*}{In-degree and $|S_{v}|$} & \multirow{2}{*}{Item-graphs} \\
    Text-BKNN-F &  & & & & \\\bottomrule
  \end{tabular}
  \caption{Result of the hyperparameters (columns) tuning when models (rows) are optimized on each validation metric (Amazon dataset). For details about the normalization methods explored, see Section \ref{sec:normalization}.}
  \label{tab:hyperparameters}
\end{table*}

\subsection{Qualitative analysis of random semantic matches} \label{sec:qualitative}
Table \ref{tab:explainability_match} provides random matches found in 5 random per-item graphs. The first two examples have lower semantic similarity than the last three examples. Both sentences of the first example are descriptive and the association focuses on the ``killing'' aspect. The second match is even unclearer; the only commonality is a mention of some female character / actress. 
The third example is made of a long and detailed tail sentence. The explicit reference to the Madea cinematic universe and the emphasized funniness are expressed in both sentences. The fourth pair of sentences refers to both the item's name they review as well as the ``family'' audience. We remark that mentions to item names or titles in review sentences may mislead the whole system, as it can trigger irrelevant semantic matches, especially if names are long. The fifth example includes a direct recommendation for lovers of suspense in both sentences.

5 random matches found in the global graph are shown in Table \ref{tab:explainability_global_match}. The first match shows a pair of sentences reviewing the same item, in which the item's name does not appear. Even if the name of the director is common to both sentences, references to the Battle of Mount Austen are phrased with different vocabulary. Indeed, the head sentence refers to ``empathy and apathy'' during ``the conflict at Guadalcanal'' while the tail sentence describes ``lush beauty of the South Pacific'' contrasting with ``the destruction of war''.
The second and third examples are made of short sentences where head and tail sentences are almost identical. These matches convey limited information when taken out of context. The last two matching pairs are triggered by common vocabulary or clauses but can't be considered as matches in users' preference.

\subsection{Domain-specific sentence embeddings} \label{domain_specific}
\begin{figure}
    \centering
      \includegraphics[width=0.9\linewidth]{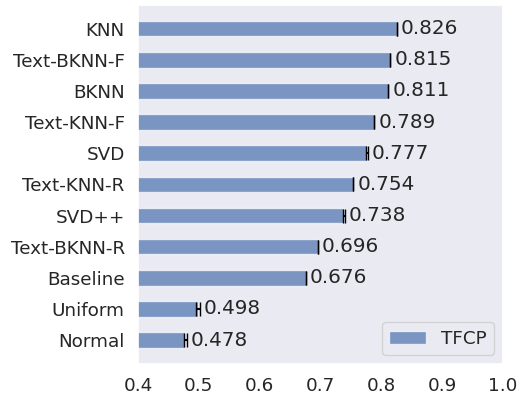}
    \caption{Bar chart of the models' test TFCP scores (average and standard deviation) on the 2018 version of the Amazon dataset, when we considered \textbf{domain-specific embeddings} rather than USE's embeddings.}
    \label{fig:raghuram}
\end{figure}

In addition to the pre-trained USE model, we experimented with domain-specific embeddings. We fine-tuned a pre-trained T5 \citep{2020t5} language model on an item prediction classification task, using the 2018 version of the Amazon Product Review dataset. We then applied our method to the representations learned by the fine-tuned T5 model, in the hope that these representations would be finer than the more general USE representations. Results of baselines and our best models (cf. Figure~\ref{fig:raghuram}) show that there is no clear advantage to using domain-specific sentence representations over general embeddings.

\begin{table*}
  \centering
  \small
  \begin{tabular}{p{2cm} c p{12.5cm}} \toprule
    \multicolumn{1}{p{2cm}}{Item} & Semantic similarity &   \multicolumn{1}{c}{Matching sentences in the train set} \\\midrule

    \multirow{4.5}{*}{\parbox{2cm}{\raggedright Spartan}} & \multirow{4.5}{*}{$0.69$} & / / Spoiler alert / / Our hero , of course , is a loner - - which is a sure tip - off that anyone he sort of becomes close to is going to get killed .\\ \cmidrule[0.0pt](l){3-3}
    & & Kilmer's character survives and even thrives in this morass because he is an unsentimental machine , the " Spartan " ideal of an all - male world where it's kill or be killed , and there's plenty of both - even a suicide .  \\\hline

    \multirow{2.5}{*}{\parbox{2cm}{\raggedright Shopgirl}} & \multirow{2.5}{*}{$0.66$} & I have a feeling there is probably more of her left on the cutting room floor .\\ \cmidrule[0.0pt](l){3-3}
    & & It's no surprise that she's on medication for depression . \\\hline

    \multirow{5.5}{*}{\parbox{2cm}{\raggedright Madea's Witness Protection}} & \multirow{5.5}{*}{$0.87$} & This movie is a very funny movie and Madea is trying to be a private Detective and she finally gets her man\\ \cmidrule[0.0pt](l){3-3}
    & & This Madea was the first time in the series to be released in the summer and did pretty well as the second highest grossing Madea film ever behind Madea Goes to Jail and this one actually crossed over with other audiences and other demographics besides his usual african - american audience and this one was the funniest film of the series and Tyler Perry's best film yet . \\\hline

    \multirow{3.5}{*}{\parbox{2cm}{\raggedright Radio}} & \multirow{3.5}{*}{$0.82$} & ' Radio ' is a wonderful film that the whole family can enjoy and learn from together !\\ \cmidrule[0.0pt](l){3-3}
    & & Radio is good , touching and very sad at times but it has a lot to learn about standing on its own as a family classic . \\\hline

    \multirow{2.5}{*}{\parbox{2cm}{\raggedright The Gift}} & \multirow{2.5}{*}{$0.75$} & Highly recommended for suspense fans .\\ \cmidrule[0.0pt](l){3-3}
    & & I would reccomend that anyone who likes mystery or suspense thrillers should go see this . \\\bottomrule

  \end{tabular}
  \caption{Random examples of semantic matches when our system considers \textbf{per-item graphs}. The last column shows the head sentence first and the tail sentence then.}
  \label{tab:explainability_match}
\end{table*}

\begin{table*}
  \centering
  \small
  \begin{tabular}{c p{14.8cm}} \toprule
    Semantic similarity &
    \multicolumn{1}{c}{Matching sentences in the train set} \\\midrule
    
    \multirow{4.5}{*}{$0.81$} & \textbf{The Thin Red Line} -- / / Director Terrence Malick focuses on the conflict at Guadalcanal from ground up and showing empathy and apathy along the way .\\ \cmidrule[0.0pt](l){2-2}
    & \textbf{The Thin Red Line} -- Mr . Malick captures the lush beauty of the South Pacific and uses it to perfectly contrast with the destruction of war .  \\\hline

    \multirow{2.5}{*}{$1.00$} & \textbf{Seabiscuit} -- This is one of my all time favorite movies .\\ \cmidrule[0.0pt](l){2-2}
    & \textbf{The Jazz Singer} -- This is one of my all - time favorite movies .  \\\hline

    \multirow{2.5}{*}{$0.86$} & \textbf{Tropic Thunder} -- it stole the show .\\ \cmidrule[0.0pt](l){2-2}
    & \textbf{My Week with Marilyn} --steals the show . \\\hline
    
    \multirow{3.5}{*}{$0.75$} & \textbf{Heist} -- It's a known fact that he uses a metronome in order to keep his dialogue to have a certain rhythm to it .\\ \cmidrule[0.0pt](l){2-2}
    & \textbf{Black Dynamite} -- He draws attention to it by repeatedly glaring at the mic throughout the scene , but doesn ' t miss a beat of the dialogue .\\\hline

    \multirow{3.5}{*}{$0.88$} & \textbf{Grown Ups} -- Now I will say from the get go this film is not for everybody and I have noticed that some people just don ' t get this film .\\ \cmidrule[0.0pt](l){2-2}
    & \textbf{Barbarella} -- This film is definately not for everyone \& I ' d honestly recommend that most people rent it before they but it .\\\bottomrule

  \end{tabular}
  \caption{Random examples of semantic matches when our system considers the \textbf{global graph}. The second column shows the head sentence first and the tail sentence then, along with the respective item they review.}
  \label{tab:explainability_global_match}
\end{table*}

\end{document}